\documentclass[final]{cvpr}

\usepackage{times}
\usepackage{epsfig}
\usepackage{graphicx}
\usepackage{amsmath}
\usepackage{amssymb}
\usepackage{lscape}

\usepackage[pagebackref=true,breaklinks=true,colorlinks,bookmarks=false]{hyperref}

\def\cvprPaperID{27} 
\def\confYear{CVPR 2021}

\begin{document}
	
	\title{Domain Adaptation and Active Learning for Fine-Grained Recognition in the Field of Biodiversity}
	
	\author{Bernd Gruner$^{1,2}$, Matthias Körschens$^1$, Björn Barz$^1$ and Joachim Denzler$^{1,2}$\\\\
		$^1$Computer Vision Group, Friedrich Schiller University Jena, Jena, Germany\\
		{\tt\small \{matthias.koerschens,bjoern.barz,joachim.denzler\}@uni-jena.de}\\
		$^2$Institute of Data Science, German Aerospace Center (DLR), Jena, Germany\\
		{\tt\small bernd.gruner@dlr.de}\\		
	}
	
	\maketitle

	\begin{abstract}
		Deep-learning methods offer unsurpassed recognition performance in a wide range of domains, including fine-grained recognition tasks. However, in most problem areas there are insufficient annotated training samples. Therefore, the topic of transfer learning respectively domain adaptation is particularly important. In this work, we investigate to what extent unsupervised domain adaptation can be used for fine-grained recognition in a biodiversity context to learn a real-world classifier based on idealized training data, e.g. preserved butterflies and plants. Moreover, we investigate the influence of different normalization layers, such as Group Normalization in combination with Weight Standardization, on the classifier. We discovered that domain adaptation works very well for fine-grained recognition and that the normalization methods have a great influence on the results. Using domain adaptation and Transferable Normalization, the accuracy of the classifier could be increased by up to 12.35~\% compared to the baseline. Furthermore, the domain adaptation system is combined with an active learning component to improve the results. We compare different active learning strategies with each other. Surprisingly, we found that more sophisticated strategies provide better results than the random selection baseline for only one of the two datasets. In this case, the distance and diversity strategy performed best. Finally, we present a problem analysis of the datasets.
	\end{abstract}
	
	\section{Introduction}
	Deep learning methods are versatile and offer outstanding recognition performance in a wide range of tasks. One example for that is the visual recognition of classes that differ only in subtle optical details, which is also known as fine-grained classification. In various competitions, like Fine-Grained Visual Categorization \footnote{\url{https://sites.google.com/view/fgvc7/home}}, these methods achieved the best results. This work deals with fine-grained image classification of butterflies and plants in the wild, which we also refer to as the target domain. The problem in this area is that we do not have annotated data for the target domain, but labeled preserved butterflies and plants, which is our source domain. Unfortunately, the classifier learned on the source domain does not give sufficient results on the data of the target domain. The data from the two domains are usually acquired under different conditions and are only a snapshot of the reality. The resulting difference is called domain shift and limits the recognition performance on the target domain.
	
	In section ~\ref{sec:relatedWork} we give an overview of several methods to bridge the domain shift. The focus of this work is on the Unsupervised Adversarial Domain Adaptation \cite{ganin2015domainadversarial} framework, as it still provides state-of-the-art performance. It reduces the domain shift by including the unlabeled data into the training process. The aim of our work is to investigate how well this method works in the area of fine-grained classification.
	
	Furthermore, normalization layers have proven to be an essential part of the network architecture, as they allow the training to converge faster and reduce the sensitivity to the initialization of parameters and hyperparameters \cite{LecunBackprop}. However, the popular method Batch Normalization \cite{ioffe2015batch} has a disadvantage in the area of domain adaptation, which affects the results negatively \cite{romijnders2018domain}. Therefore, we test and compare Batch Normalization against the alternatives, which are Domain Agnostic Normalization \cite{romijnders2018domain}, Transferable Normalization \cite{NIPS2019_8470} and Group Normalization \cite{wu2018group} in combination with Weight Standardization \cite{qiao2019weight}. The latter is already successfully applied in transfer learning \cite{alex2019big}, but to the best of our knowledge not yet tested in a domain adaptation scenario.
	
	To further improve the results of the classifier, we added an active learning component, which helps us to select target samples for annotation in a structured way. In our experimental setup, we only simulate the annotation process as we already have all the labels for the target dataset in order to enable an evaluation. This can be regarded in contrast to the real application scenario where the annotation is carried out by a domain expert and is much more time-consuming. 
	
	We investigate how many target samples have to be annotated randomly in order to achieve a result comparable to fully supervised learning on the target domain. Further, this baseline experiment is expanded by comparing it to various active learning strategies: Expected Model Output Change \cite{kding2016active}, Distance and Diversity \cite{10.55553041838.3041846}, Importance Weight \cite{su2019active} and Certainty sampling. To the best of our knowledge, Expected Model Output Change has not yet been used in the field of domain adaptation.
	
	\section{Related Work}
	\label{sec:relatedWork}
	There are various ways to deal with a domain shift, but it is particularly difficult when there are no or only a few annotated data points from the target domain. Therefore, the simplest solution is to collect more labeled target data, which is in most cases very expensive and time-consuming. Other possibilities are the use of improved methods for feature extraction, further data augmentation \cite{volpi2018generalizing, choi2019selfensembling}, Mix Match \cite{berthelot2019mixmatch, rukhovich2019mixmatch}, semi-supervised methods \cite{saito2019semisupervised, mishra2021surprisingly}, domain adaptation \cite{36364, zhu2017unpaired}, etc. We will focus on the latter in this work.
	\subsection{Domain Adaptation}
	The domain adaptation approach tries to adjust the domains so that the learned classifier can generalize over different domains. This can be done either in the pixel space by aligning the data points before the actual processing or in the representation space by aligning the representations during feature extraction. In the following, we present a selection of approaches based on different principles and rest on the surveys \cite{wilson2018survey} and \cite{wang2018deep}.
	
	Divergence-based approaches try to minimize some divergence criterion between the distributions of the two datasets. For this reason, it leads to a domain invariant feature representation and the classifier's ability to generalize well over the domains. Criteria for measuring the distance for example are Maximum Mean Discrepancy (MMD) \cite{Rozantsev_2019}, Contrastive Domain Discrepancy (CCD) \cite{sun2016deep}, Correlation Alignment (CORAL) \cite{kang2019contrastive} or the Wasserstein metric \cite{damodaran2018deepjdot}.
	
	Other approaches are adversarial-based, which refers to any method that utilizes an adversary or an adversarial process during training. Some of the solutions here are based on Generative Adversarial Networks (GANs) \cite{goodfellow2014generative}, as Coupled Generative Adversarial Networks (CoGAN) \cite{liu2016coupled}. This approach generates synthetic target data for training. A further procedure is presented in \cite{yoo2016pixellevel}, which uses a source-target converter network architecture. The Domain Adversarial Neural Network (DANN) \cite{ganin2015domainadversarial}, which is built on the Unsupervised Adversarial Domain Adaptation (UADA) framework, uses a discriminator for the domain classification and tries to optimize the feature extractor by a confusion loss so that the resulting representation is domain invariant.
	
	In the reconstruction-based approaches, shared representations for the domains are created by an auxiliary reconstruction task. This is applied in the Deep Reconstruction Classification Network (DRCN) \cite{ghifary2016deep}, which classifies the source data and reconstructs the unlabeled target data. The network learns to transform source data into data that is similar to the target dataset, while creating a common representation. Another possibility to use domain adaptation is in Cycle GANs \cite{zhu2017unpaired}. They can learn to convert data from one domain to another, which is achieved by training two GANs simultaneously. Furthermore, Conditional GANs \cite{isola2016imagetoimage} can be applied to transfer from one domain to another by using a simple encoder-decoder architecture.
	
	For the domain adaptation in this work the UADA approach is used, as it provides still state-of-the-art performance and can be used for unsupervised and semi-supervised domain adaptation.
	\subsection{Normalization}
	Normalization layers have become an important part of the network architecture because of their advantages during the training process (see subsection \ref{subsec:norm}). However, Batch Normalization \cite{ioffe2015batch} brings some problems by introducing dependencies between the different domains when using multi-domain batches \cite{romijnders2018domain}. Some further developments are addressing this weakness. Therefore, Group Normalization \cite{wu2018group} in combination with Weight Standardization~\cite{qiao2019weight}, no longer uses batch statistics for normalization. The Domain Agnostic Normalization \cite{romijnders2018domain} tries to counteract the weakness by using only the source domain's statistics. The Domain-Specific Batch Normalization \cite{chang2019domainspecific} solves the dependencies by using the statistics of the respective domain for normalization. Another possibility is to use Adaptive Batch Normalization \cite{LI2018109}, which modulates the statistics from the source domain to the target domain. Furthermore, there is the Transferable Normalization \cite{NIPS2019_8470}, which includes a transferability factor besides the independent statistics. 
	
	Group Normalization in combination with Weight Standardization is the only method mentioned that has not been specially developed for domain adaptation and is not evaluated in this field. We test and compare Group Normalization in combination with Weight Standardization, Transferable Normalization and Domain Agnostic Normalization, because they all outperform Batch Normalization in their application, but they are not compared with each other in their publication.
	\subsection{Active Learning}
	Active learning is a structured approach to select samples from unlabeled data using a selection strategy. Afterward, these samples are annotated. There are different strategies, for example criteria based on uncertainty \cite{yang2017active, LEWIS1994148}, reducing expected error \cite{10.5555645530.655646, Zhu03combiningactive}, diversity \cite{7780682} or maximizing expected label changes \cite{inproceedingsemoc, 7299063}. Further combinations of these criteria have been developed such as Importance Weight \cite{su2019active} or Distance and Diversity \cite{10.55553041838.3041846}.
	
	In previous research, the combination of domain adaptation and active learning methods was investigated. This is accomplished in Active Learning Domain Adapted (ALDA) \cite{10.5555/1860625.1860629, 10.55552034161.2034169} for sentiment and landmine classification and in Active Adversarial Domain Adaptation (AADA) \cite{su2019active} for vision tasks like object detection.
	
	In this work, we also combine domain adaptation with active learning but with a focus on fine-grained classification in the field of biodiversity. Hence, we test and evaluate different selection strategies, including the Expected Model Output Change strategy, which has not been evaluated in this field before. 
	
	\section{Methods}
	In the following, we will briefly review the domain adaptation, normalization, and active learning methods, whose feasibility for a fine-grained recognition scenario we are going to evaluate. The core of the system is the framework Unsupervised Adversarial Domain Adaptation \cite{ganin2015domainadversarial}, which we present in the first section of this chapter. Then we explain the different normalization methods and finally describe the active learning approach by which the system has been extended.
	\subsection{Unsupervised Adversarial Domain Adaptation}
	The Unsupervised Adversarial Domain Adaptation (UADA) framework deals with the scenario where no labels are available in the target domain. It is based on Ben-David's theory about domain adaptation~\cite{NIPS2006_2983, 36364}, which addresses the problem that the source and target data come from two different datasets, thus from two different distributions. Hence, the classification error on the target data may not be close to that on the source data. A solution developed in the paper of Ben-David et al.~\cite{36364} is to align the representations of the source and target domain so that an algorithm can no longer distinguish between them. 
	In the UADA framework a Convolutional Neural Network (CNN) is used as a feature extractor $G_{F}$, which in our case is a ResNet-50~\cite{he2015deep} pretrained on ImageNet~\cite{ILSVRC15}. These features are used in the label predictor $G_{L}$ for classification. This classical structure is extended by a so-called domain discriminator $G_{D}$, which uses the features to predict the domain of the representation. For the end-to-end training, a cross-entropy classification loss is used to train the feature extractor and the label predictor. Furthermore, the domain discriminator is trained with a binary cross-entropy loss. Additionally, a confusion loss $L_{conf}$ is used in the feature extractor, which is defined in \cite{tzeng2015simultaneous} as follows:
	\begin{equation}
	\label{equ:targetconfloss}
	L_{conf} = -\sum_{x \in T} \log(1-G_{D}(G_{F}(x))),
	\end{equation} 
	where $T$ stands for all target data points and $G_{D}(G_{F}(x))$ for the output of the domain discriminator when the feature representation of $x$ is passed in.
	It should result in the representations being more aligned so that the domain discriminator can no longer distinguish between the domains.  
	\subsection{Normalization Methods}
	\label{subsec:norm}
	In deep neural networks with many hidden layers, the use of normalization layers is standard. It has the positive effect that the training converges faster and reduces the sensitivity to the initialization of parameters and hyperparameters \cite{NIPS2005_2920, LecunBackprop}. A major disadvantage concerning domain adaptation is that the popular Batch Normalization (BN) \cite{ioffe2015batch} is based on statistics that depend on the complete batch. Since it is common for domain adaptation approaches that a batch contains data from a source and target domain, the normalization of an exclusive representation also depends on the representations of the other domain. This is a problem concerning the unlabeled target domain because these representations get no supervision from a prediction only from other images via the calculated statistics, which results in havoc on the target representation \cite{romijnders2018domain}. Some further developments address and resolve these weaknesses. Domain Agnostic Normalization (DAN) \cite{romijnders2018domain} is a very recent technique and has achieved excellent results in image segmentation. It has outperformed Batch Normalization in the domain adaptation task in the mentioned publication by 8.6~\% accuracy. Therefore, this method is investigated for our fine-grained recognition task. It overcomes the problem of Batch Normalization by resolving the dependency between the different domains using only the statistics of the source domain for normalization. 
	
	A similar approach is followed by Transferable Normalization (TransNorm) \cite{NIPS2019_8470}, which is also a promising new approach. It eliminates the dependency between the domains and weights the channels according to transferability. In the paper it also improved the accuracy by up to 10~\% compared to Batch Normalization depending on the dataset used.
	
	Group Normalization \cite{wu2018group} in combination with Weight Standardization (GN + WS) \cite{qiao2019weight} is another alternative used, which, unlike the other two, was not specifically designed for domain adaptation and was not applied in this area before. It performs a batch-independent normalization of a group of channels. From the papers, we know that it has outperformed Batch Normalization by 0.79~\% accuracy in the ImageNet classification task. Furthermore, this good performance is confirmed for transfer learning in \cite{alex2019big}.
	\subsection{Active Learning}
	In the case of unsupervised domain adaptation, we have a labeled source and an unlabeled target dataset. It is impractical to label the complete target dataset for a small project as it would be too time-consuming and costly. However, it is possible to annotate a part of it, and the active learning approach helps us to select informative samples. In our experimental setup, we have the labels for the target domain to be able to evaluate the results of the system, in contrast to a real-world application.
	
	Batch-wise active learning with a batch size of $k$ is used, because otherwise, the network would need to be trained completely from scratch again for every newly annotated example, which would be too resource-intensive.
	For the selection of a suitable sample for annotation, a criterion is needed. There are many criteria, and they all try to find the most informative data point based on various principles. 
	Random selection is used as a baseline in our study. The advantage of this method is that it consumes very few resources and is independent of the size of the unlabeled dataset. 
	
	A simple and resource-saving method is Certainty (Cer) sampling. It uses entropy to select the sample where the network is most certain. The aim is to avoid errors in the prediction in case of high scores.
	
	A more sophisticated strategy is Distance and Diversity (DivDis) \cite{10.55553041838.3041846}. It is a combination of the distance to the decision boundary, which is a kind of uncertainty measure of the classifier, and the diversity. Therefore, we use the cosine similarity. This method, first developed in 2003, is a combination of criteria and has achieved excellent results in the field of Support Vector Machines in the mentioned paper. The approach is then transferred to Deep Neural Network-based classifiers.
	
	Importance Weight (IWERM) \cite{su2019active} combines the importance of an example with the uncertainty. This recent strategy was developed for the adversarial domain adaptation approach and uses the discriminator to calculate the importance.
	
	The Expected Model Output Change (EMOC) \cite{kding2016active} is a more sophisticated strategy and uses a higher amount of computational resources. It tries to find the sample that causes the greatest possible change in the network output after the retraining. EMOC has not yet been used in the field of domain adaptation.
	
	\section{Datasets}
	In this section, we introduce our self-created datasets used in this study. They consist of already existing datasets, which were enriched with samples from the Global Biodiversity Information Facility (GBIF) \footnote{\url{https://www.gbif.org}}. We downloaded images from the GBIF database and manually rechecked them and sorted out images without the expected object. The two datasets are created for the fine-grained recognition of plants and butterflies and each of them has two sub-datasets. The first subset used is the source dataset which contains images from the source domain, the second subset is the target dataset which contains images from the target domain. Both subsets consist of the same classes hence there is no category shift. These two datasets are split into 72~\% training, 8~\% validation, and 20~\% test data. The following two subsections explain the specifics of each dataset.
	\subsection{Butterfly Dataset}
\begin{figure}[t]
		\centering
		\includegraphics[width=0.9\linewidth]{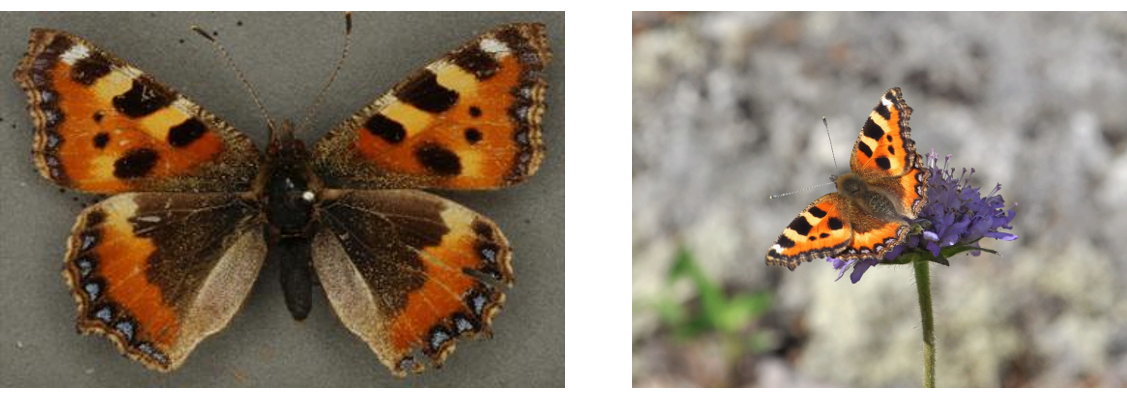}
		\caption[Examples from the butterfly dataset]{These are two samples of the butterfly \textit{Algais Urticae - Small Tortoiseshell} . In the left column a specimen from the source domain can be seen and on the right side, there is the corresponding image from the target domain.}
		\label{fig:Bfds}
	\end{figure}

	The butterfly dataset includes 15 different species of butterflies. Each of these classes contains between 150 to 440 images. While the target domain consists of observations of wild butterflies, the source domain is preserved specimens of butterflies. Some examples are illustrated in figure \ref{fig:Bfds}. In the source domain, the butterflies are always photographed from above at almost the same angle. Furthermore, they are completely visible in the pictures and the conservation process has not changed the colors. In the target domain, the butterflies are photographed from different angles and distances. The background varies considerably in the target domain compared to the source domain. In summary, a certain shift between the domains can be observed.
	
	The dataset is a combination of the UCL Butterflies \cite{JohnsCVPR2015} and Leeds Butterflies \cite{Wang09} datasets enriched with samples from the GBIF database. 
	\subsection{Plant Dataset}
	The plant dataset includes 11 different plant species with 37 images per class. This dataset is based on the Herbarium Challenge dataset \cite{tan2019herbarium} with 680 species and over 46000 images, but they are not uniformly distributed. There are only a few classes with over 40 images. The target subset is downloaded from the GBIF database and contains only a few examples for each plant. We have selected the overlapping classes, which have at least 37 images so that we have a balanced source and target dataset. 
	
	\begin{figure}[t]
		\centering
		\includegraphics[width=0.9\linewidth]{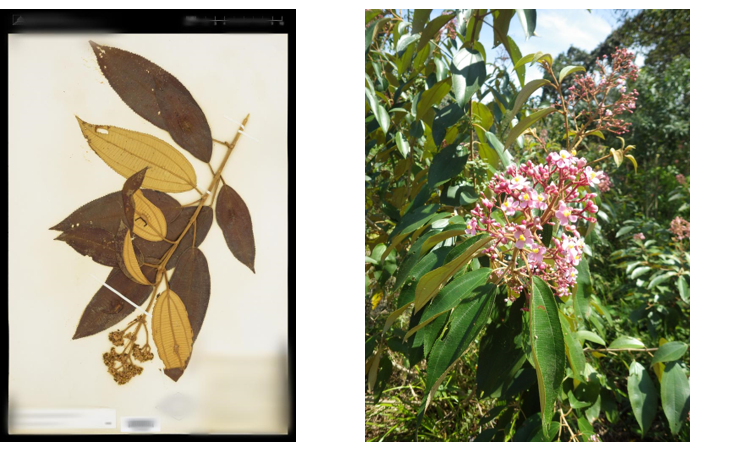}
		\caption[Examples from the plant dataset]{These are two samples of the plant \textit{Conostegia xalapensis (Bonpl.) D.Don ex DC}. In the left column a specimen from the source domain can be seen and on the right side, there is the corresponding image from the target domain.}
		\label{fig:Pds}
	\end{figure}
	In figure \ref{fig:Pds}, some examples are shown. Preserved specimens are the constituents of the source domain. The classes of plants range from flowers to bushes and trees. Hence, the sizes are also completely different. Often, only the leaves can be seen in the images. The conservation process changes the shape of the flowers and their color is lost. In the target domain, which shows the plants in the wild, key parts of the plants, such as flowers, are photographed and in some cases, even the whole plant can be seen on the image. Usually, the leaves of the plant are not in focus. It can be summarized that the domains are very different and it is difficult to find common features. 
	\section{Experimental Results and Evaluation}
	In the beginning, we introduce the baseline experiments that are used to rank the performance of the domain adaptation system. This is followed by the presentation of the domain adaptation experiments with different normalization methods. We perform a hyperparameter optimization of learning rate and L2-regularization using grid search. The results of the grid search for all setups can be found in the supplementary materials A.1 and A.2.  In general, we ran all experiments three times for 100 epochs with a batch size of 32 and report the averaged mean accuracy per class. In preprocessing, random flip and scale augmentation are used as data augmentation \cite{he2015deep} and the images are scaled to an input size of 224 x 224.
	
	In a further experiment, the combination of active learning and domain adaptation is examined. Therefore, a baseline for the active learning approach is created in the first part of the experiment. The second part investigates and compares different active learning strategies. Finally, we analyze existing problems with the datasets.
	\subsection{Baseline: CNN-Classifier}
	\label{sec:experiment1}
	In order to determine whether the unsupervised domain adaptation approach works in the field of fine-grained recognition and how well it performs, we create a baseline. This is done by using a CNN-classifier in two different setups and thus determining an upper bound and a baseline. Therefore, only the feature extractor and the label predictor are used from the domain adaptation approach, which corresponds to a CNN-classifier architecture. For a baseline (source-only), it is trained with the annotated source data and evaluated on the target test data. The upper bound (target-only) is created by training and testing on the annotated target data.
	
	Table~\ref{tab:babut} shows an overview of the results when training on the butterfly dataset. The baseline achieves an accuracy of 67.52~\% when evaluated on the target test set.  Comparing this with the result of the target-only classifier, it is evident that there is a gap of 27.07~\% and this is likely caused by the domain shift. In general, the result of the target-only classifier confirms that the fine-grained classification task can be completed with an accuracy of 94.59~\%. 
	
	\begin{table}
		\begin{center}
			\begin{tabular}{|c|c|c|}
				\hline 
				Evaluation / Training  & Source Domain & Target Domain \\ 
				\hline 
				Source Domain & 100 \% & 89.04 \% \\ 
				\hline 
				Target Domain & 67.52 \% &  94.59 \% \\ 
				\hline 
			\end{tabular}
			\vspace*{0.4mm}
			\caption{In this table, the accuracy of the classifier (source-only, baseline) trained on the source butterfly dataset and the classifier (target-only, upper bound) trained on the target butterfly dataset are shown.}
			\label{tab:babut}
		\end{center}
	\end{table} 
	\begin{table}
		\begin{center}
			\begin{tabular}{|c||c|c|}
				\hline 
				Evaluation / Training  & Source Domain & Target Domain \\ 
				\hline \hline
				Source Domain & 84.13 \%  & 18.88 \% \\ 
				\hline 
				Target Domain & 15.04 \%\% &  81.23 \% \\ 
				\hline 
			\end{tabular}
			\vspace*{0.4mm}
			\caption{In this table, the accuracy of the classifier (source-only, baseline) trained on the source plant dataset and the classifier (target-only, upper bound) trained on the target plant dataset are shown.}
			\label{tab:bapl}
		\end{center} 
	\end{table} 
	Furthermore, in table~\ref{tab:bapl}, we present the results of the source-only classifier on the plant dataset. There is a clear difference to the results of the training on the butterfly dataset. The few images per class likely cause that the classifier cannot generalize well enough to overcome the large shift between domains. The best result achieved here is 15.04~\%, which is, nevertheless, slightly better than a random guess. In contrast, the target-only classifier achieves a result of 81.23~\% accuracy. 
	\begin{figure*}[t]
		\centering
		\includegraphics[scale=0.4]{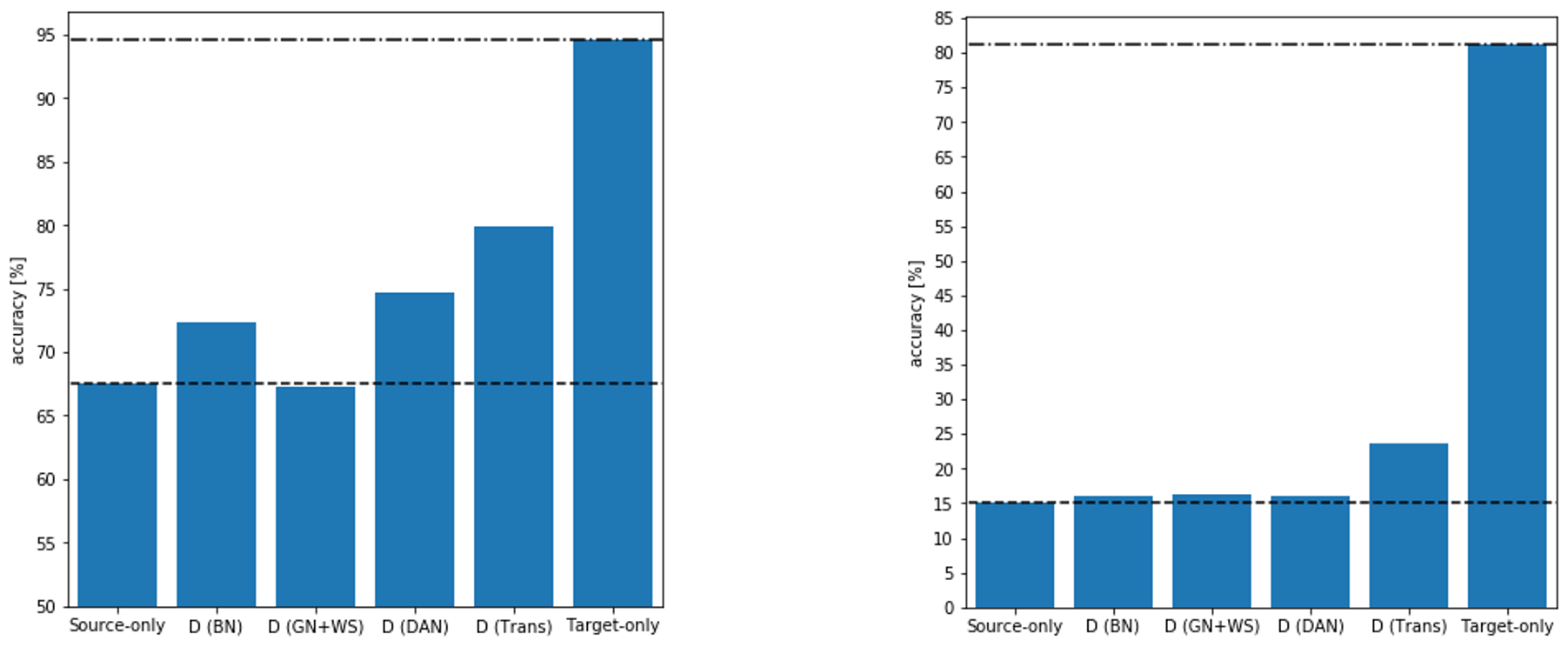}
		\caption{A comparison between the results (source-only and target-only) from the experiments in section \ref{sec:experiment1} and the domain adaptation approach is shown. The diagram on the left shows results for the butterfly dataset and the one on the right for the plant dataset. The domain adaptation setup is divided into the different normalization methods used: Batch Normalization - D (BN), Group Normalization in combination with Weight Standardization - D (GN+WS), Domain Agnostic Normalization - D (DAN) and Transferable Normalization - D (Trans).}
		\label{fig:com_bas_but}
	\end{figure*} 
	There is a domain shift in both datasets and great potential when comparing the results of the source- and target-only classifier, which should be used by the domain adaptation approach in the following experiment. 
	\subsection{Domain Adaptation Approach}
	\label{sec:experiment2}
	This experiment aims to determine whether and how well the unsupervised domain adaptation approach works for our fine-grained recognition tasks. We test four setups of the domain adaptation architecture with different normalization methods. For the training, annotated source and non-annotated target data are used. Afterward, the evaluation is performed on the target test data. 
	
	In figure~\ref{fig:com_bas_but} on the left side, we present the results of the domain adaptation approach for the butterfly dataset. The upper bound and baseline from the experiments in section~\ref{sec:experiment1} are used for comparison. In between, there are the domain adaptation architectures with the different normalization methods. It has to be noted that the GN+WS \cite{wu2018group, qiao2019weight} normalization performs worse than BN \cite{ioffe2015batch} and does not even offer any improvement to the source-only classifier. The other three normalization methods show an improvement compared to the baseline. TransNorm \cite{NIPS2019_8470} performs best with 79.87~\% accuracy. This confirms for the butterfly dataset that the domain adaptation for the fine-grained recognition task works. However, between the results of the domain adaptation and the upper bound, there is still a gap of 14.72~\% accuracy. In the upcoming experiment, we try to reduce this gap by using active learning methods. 
	
	On the right side in figure \ref{fig:com_bas_but} an overview of the experiments on the plant dataset is provided. The results from section~\ref{sec:experiment1} are also used here for comparison. It becomes clear that this time only the TransNorm shows an advantage over the baseline. BN, GN+WS and DAN \cite{romijnders2018domain} do not provide an improvement. The reason could be the small dataset and the difficult classification task with the large domain shift. We discuss the specific difficulties with the dataset in more detail in section \ref{sec:DisProb}. TransNorm can achieve a substantially better result of 23.54~\% accuracy by weighting the different channels according to their transferability. However, the gap between this result and the one of the target-only classifier is with 57.69~\% accuracy huge, compared to the gap in the corresponding experiment with the butterfly dataset. 
	
	In summary, it can be said that the normalization layers have a substantial influence on the classification result of the domain adaptation classifier. When using the TransNorm, the domain adaptation approach works well in the field of fine-grained recognition of butterflies and plants. This classifier outperformed the source-only training up to 12.35~\% accuracy only by incorporating the unlabeled target data into the training process.
	Despite the aforementioned weakness, BN offered better results than the baseline. In contrast, the classifier with GN+WS only achieved results comparable to the source-only classifier. To the best of our knowledge, the latter method was applied for the first time in this area. However, it is not recommended for the datasets used in this work. 
	\subsection{Domain Adaptation combined with Active Learning}
	\label{sec:experiment3}
	To further improve the results of the domain adaptation classifier, parts of the target dataset can be annotated using active learning methods. Firstly, we determine the number of target samples that have to be annotated to obtain a comparable result to the upper bound from the experiment in section \ref{sec:experiment1}. Therefore, random selection is used as a baseline and the determined hyperparameters from section \ref{sec:experiment2} are applied. For both datasets, ten images per round are annotated and then the network is trained from scratch.
	\begin{figure}[t]
		\centering
		\includegraphics[width=1.0\linewidth]{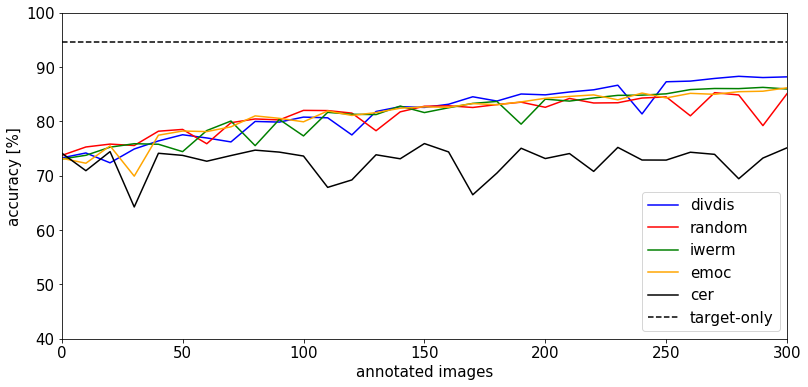}
		\caption[Comparison of different active learning strategies (butterfly dataset, 0-300 images)]{In the diagram, the five different active learning strategies, which are color-coded, are compared. For this purpose, ten images per round have been annotated from the butterfly target dataset over 30 rounds. }
		\label{fig:comp_al_str_but}
	\end{figure} \noindent
	Within 30 rounds, 300 images are annotated for both datasets. In the case of the plant dataset, this is already the complete training set. When using the butterfly dataset, we set $k=500$ after the first 30 rounds. Thus, the active learning approach can be applied with an acceptable amount of resources.
	
	In the second part of the experiment, we use the same setup to test and compare more sophisticated selection strategies as a key component of the active learning approach. The strategies EMOC \cite{kding2016active}, DivDis \cite{10.55553041838.3041846}, Cer and IWERM \cite{su2019active} are compared with the random selection tested in the first part. Furthermore, a second setup is used, where we increase the number of runs to 17 and reduce the number of rounds to ten. Thus, a total of 100 images are annotated. This setup was only carried out for the plant dataset in order to identify possible differences between the strategies.\\

	Figure~\ref{fig:comp_al_str_but} shows a plot that illustrates the results of the training on the butterfly dataset for the first 30 rounds. The random selection baseline, which is shown in red, increased by 10~\% after 300 annotations. A clear upward trend could be seen in all curves as well, except that of the Cer strategy. They are also not strictly monotonously increasing. After some rounds, they are even decreasing. There are some outliers in the curves, which are not averaged out due to the low number of repetitions. In general, the plot clearly shows that all results of the strategies are very close to each other except for Cer. The Cer strategy drops far below the others after the first 50 annotated images and the result after 300 annotated images is only about one percent better. It can also be seen that the DivDis strategy gives the best results from about 200 annotated images onwards, whereas the random strategy lies below DivDis, IWERM and EMOC.
	
\begin{figure}[t]
		\centering
		\includegraphics[width=1.0\linewidth]{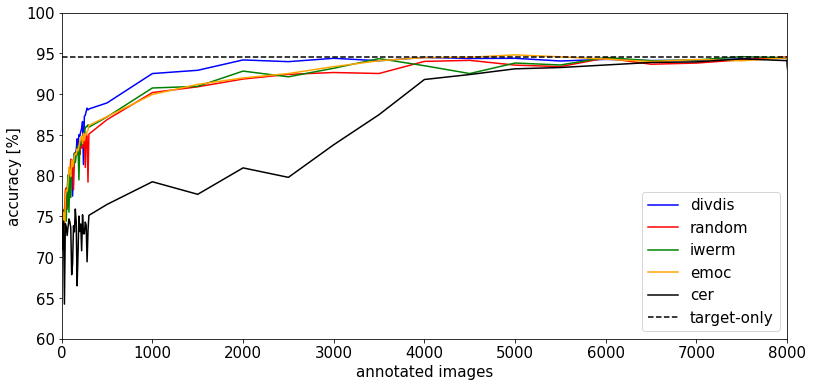}
		\caption[Comparison: active learning vs. target-only (butterfly dataset, 0-5000 images)]{This plot presents the results of active learning with five different active learning strategies on the butterfly dataset compared to the result of the target-only training (black, dashed) from the experiment in section \ref{sec:experiment1}. There are 16 rounds, and 500 images are annotated in each round. The first 30 data points are taken from figure~\ref{fig:comp_al_str_but}.}
		\label{fig:comp_al_bas_big_but}
	\end{figure}

\begin{figure}[t]
		\centering
		\includegraphics[width=1.0\linewidth]{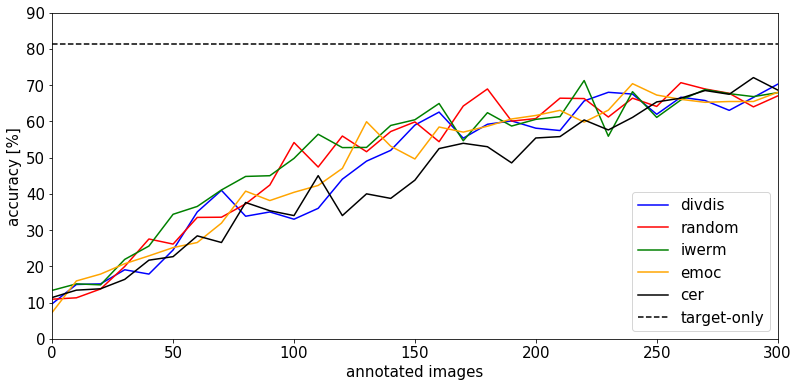}
		\caption[Comparison of different active learning strategies (plant dataset, 0-300 images)]{In the plot, the five different active learning strategies, which are color-coded, are compared. For this purpose, ten images per round have been annotated from the plant target dataset over 30 rounds. }
		\label{fig:comp_al_str_pla}
	\end{figure}

	As the accuracies are more than 10~\% below the upper bound (dashed line), we have increased the number of images to be annotated per round to 500. The results of this setup are shown in figure~\ref{fig:comp_al_bas_big_but}. The baseline (red curve) needs about 4000 annotations to achieve a comparable result to the upper bound, which corresponds to half of the target training set. The Cer strategy performs much worse with about 5000 images to annotate. The curves of the IWERM and EMOC strategies are slightly above the baseline and require about 3500 annotations. The best strategy is DivDis, which requires the fewest annotations with about 2000.\\\\	
	When using the plant dataset, ten images are annotated step by step until the complete target training set is labeled. The curve rises steepest for the first hundred annotated images that were added and then flattens further and further, as seen at the red line in figure \ref{fig:comp_al_str_pla}. The accuracy is increased by 64.5~\% over all 30 rounds but does not reach the results of target-only training even after the completed annotation of the target training set. This problem can be explained by the fact that the large domain shift means that the features have to be aligned more, so that the classifier can generalize better. However, this does not lead to a high performance in a specific domain. This can be seen when comparing the results on the other domains. The target-only classifier gives very poor results of 18~\% accuracy on the source test data compared to the domain adaptation classifier on the source data which gives very good results of 83~\% accuracy. 
	
	Looking at the other selection strategies, all curves show a clear upward trend with some outliers. Up to the first 50 annotated target samples, all strategies have similar results. Then the Cer strategy drops and only after 250 annotations, a comparable result to the others is reached again. In general, there is no strategy that can be regarded as being definitely better. In most cases the IWERM and random selection curves show the best results.
	
	In order to obtain a more stable mean value, the experiment was performed 14 additional times on the plant dataset. Due to resource restrictions, the number of rounds was reduced to ten. We still observe that the strategies barely differ. However, the 17 rounds make it possible to perform a meaningful significance test for the results. Therefore, we tested the results of every combination of two strategies for significance, using student's t-test \cite{10.2307/2331554}. Two results differ significantly if the p-value is smaller than the threshold of 0.05. The test shows that none of the strategies differ significantly from each other over all rounds. Between the strategies IWERM~\&~Cer and random~\&~Cer, the p-values of three results after different rounds show that Cer is significantly worse than IWERM and the random selection. Furthermore, the random selection also performs significantly better than the DivDis strategy in three results. In all other cases, only individual values are classified as significantly different. An overview of all p-values is presented in the supplementary materials~C.
	
	In summary, when training on the butterfly dataset, only the Cer strategy was worse than the baseline. The EMOC and IWERM strategies performed slightly better than baseline and the DivDis strategy works best, requiring only half the annotations of the baseline. Using the plant dataset, the results could be increased by up to 55~\%, but even after the complete training data annotation, no comparable result to the target-only training was reached. Even the use of more sophisticated selection strategies did not lead to significantly better results. This is probably due to the massive domain shift and the small dataset. For the plant dataset we recommend the random selection due to its resource-friendliness. 
	\subsection{Discussion of Dataset Problems}
	\label{sec:DisProb}
	In this section, we discuss possible difficulties that occur with the datasets.
	The results of the classifiers trained on the butterfly dataset are very good, despite the differences between the domains. Figure \ref{fig:Bfds} shows one example from the source domain and one from the target domain. It is immediately noticeable that the preserved specimen is huge and the patterns on the wings are clearly visible, when contrasting them to the images from the target dataset, where the butterflies are tiny. This can be challenging for the CNN, as it generally does not produce scale-invariant features \cite{Goodfellow-et-al-2016}.
	
	Furthermore, some butterflies are captured in flight and are therefore only visible out of focus. In the images it can be seen that butterflies are also photographed from different angles and sometimes the pattern on the wing is not visible. Additionally, the undersides of the wings look different from the topsides. This can also lead to recognition problems. However, the dataset contains enough images to overcome the obstacles most of the time. A collection of more examples from the dataset can be found in the supplementary materials B.1.
	
	The domain adaptation classifier trained on the plant dataset provides only minor improvements over the baseline. On the left-hand side in figure~\ref{fig:Pds}, an image from the source domain is shown and on the right-hand side, there is an example from the target domain. In general, it can be seen that the preserved specimens mostly show the leaves and only occasionally the flowers. Besides, the flowers have a different shape due to the conservation and the colors are also different.	The images of plants in the wild often show the flowers. However, there are also examples, where the complete plant is pictured. Usually the leaves are not in focus in these images and often, they can only be found very small in the background. Another point is that the source images sometimes contain only fragments of the plant. Some more examples of difficult images can be found in the supplementary materials B.2.
	
	These reasons clarify the enormous difference between the domains and show how little information can only be transferred and used for classification. The low number of training images makes it even more difficult.
	
	\section{Conclusion}
	In this work, we have shown that Unsupervised Adversarial Domain Adaptation works for fine-grained recognition in the field of biodiversity. To do so, we created two datasets of butterflies and plants. Our experiments demonstrated that the domain adaptation worked for our fine-grained recognition task and that the normalization method has a great influence on the result of the classifier. Among the methods that have been investigated, we found that GN+WS is similar to the baseline, DAN and BN are in the middle and TransNorm provides the best results. Moreover, we extended the classifier with an active learning component and compared different selection strategies with each other. In general, the Cer strategy performed worst on both datasets. For the butterfly dataset we recommend using \mbox{DivDis} as it gives the best results. When training on the plant dataset, none of the tested strategies could significantly outperform the baseline, therefore the random selection is advised.	
	
	{\small
		\bibliographystyle{ieee_fullname}
		\bibliography{paper}

\begin{thebibliography}{10}\itemsep=-1pt

\bibitem{36364}
Shai Ben-David, John Blitzer, Koby Crammer, Alex Kulesza, Fernando Pereira, and
  Jennifer Vaughan.
\newblock A theory of learning from different domains.
\newblock {\em Machine Learning}, 79:151--175, 2010.

\bibitem{NIPS2006_2983}
Shai Ben-David, John Blitzer, Koby Crammer, and Fernando Pereira.
\newblock Analysis of representations for domain adaptation.
\newblock In B. Schoelkopf, J.~C. Platt, and T. Hoffman, editors, {\em Advances
  in Neural Information Processing Systems 19}, pages 137--144. MIT Press,
  2007.

\bibitem{berthelot2019mixmatch}
David Berthelot, Nicholas Carlini, Ian Goodfellow, Nicolas Papernot, Avital
  Oliver, and Colin Raffel.
\newblock Mixmatch: A holistic approach to semi-supervised learning, 2019.

\bibitem{10.55553041838.3041846}
Klaus Brinker.
\newblock Incorporating diversity in active learning with support vector
  machines.
\newblock In {\em Proceedings of the Twentieth International Conference on
  International Conference on Machine Learning}, ICML03, pages 59--66. AAAI
  Press, 2003.

\bibitem{chang2019domainspecific}
Woong-Gi Chang, Tackgeun You, Seonguk Seo, Suha Kwak, and Bohyung Han.
\newblock Domain-specific batch normalization for unsupervised domain
  adaptation.
\newblock In {\em Proceedings of the IEEE/CVF Conference on Computer Vision and
  Pattern Recognition (CVPR)}, June 2019.

\bibitem{choi2019selfensembling}
Jaehoon Choi, Taekyung Kim, and Changick Kim.
\newblock Self-ensembling with gan-based data augmentation for domain
  adaptation in semantic segmentation.
\newblock In {\em Proceedings of the IEEE/CVF International Conference on
  Computer Vision (ICCV)}, October 2019.

\bibitem{NIPS2005_2920}
Koby Crammer, Michael Kearns, and Jennifer Wortman.
\newblock Learning from data of variable quality.
\newblock In Y. Weiss, B. Schoelkopf, and J.~C. Platt, editors, {\em Advances
  in Neural Information Processing Systems 18}, pages 219--226. MIT Press,
  2006.

\bibitem{damodaran2018deepjdot}
Bharath~Bhushan Damodaran, Benjamin Kellenberger, Remi Flamary, Devis Tuia, and
  Nicolas Courty.
\newblock Deepjdot: Deep joint distribution optimal transport for unsupervised
  domain adaptation.
\newblock In {\em Proceedings of the European Conference on Computer Vision
  (ECCV)}, September 2018.

\bibitem{inproceedingsemoc}
Alexander Freytag, Erik Rodner, and Joachim Denzler.
\newblock Selecting influential examples: Active learning with expected model
  output changes.
\newblock In David Fleet, Tomas Pajdla, Bernt Schiele, and Tinne Tuytelaars,
  editors, {\em Computer Vision -- ECCV 2014}, pages 562--577, Cham, 2014.
  Springer International Publishing.

\bibitem{ganin2015domainadversarial}
Yaroslav Ganin, Evgeniya Ustinova, Hana Ajakan, Pascal Germain, Hugo
  Larochelle, Fran{\c{c}}ois Laviolette, Mario Marchand, and Victor Lempitsky.
\newblock Domain-adversarial training of neural networks.
\newblock {\em The journal of machine learning research}, 17(1):2096--2030,
  2016.

\bibitem{ghifary2016deep}
Muhammad Ghifary, W~Bastiaan Kleijn, Mengjie Zhang, David Balduzzi, and Wen Li.
\newblock Deep reconstruction-classification networks for unsupervised domain
  adaptation.
\newblock In {\em European Conference on Computer Vision}, pages 597--613.
  Springer, 2016.

\bibitem{Goodfellow-et-al-2016}
Ian Goodfellow, Yoshua Bengio, and Aaron Courville.
\newblock {\em Deep Learning}.
\newblock MIT Press, 2016.

\bibitem{goodfellow2014generative}
Ian~J Goodfellow, Jean Pouget-Abadie, Mehdi Mirza, Bing Xu, David Warde-Farley,
  Sherjil Ozair, Aaron~C Courville, and Yoshua Bengio.
\newblock Generative adversarial nets.
\newblock In {\em NIPS}, 2014.

\bibitem{he2015deep}
Kaiming He, Xiangyu Zhang, Shaoqing Ren, and Jian Sun.
\newblock Deep residual learning for image recognition.
\newblock In {\em Proceedings of the IEEE conference on computer vision and
  pattern recognition}, pages 770--778, 2016.

\bibitem{ioffe2015batch}
Sergey Ioffe and Christian Szegedy.
\newblock Batch normalization: Accelerating deep network training by reducing
  internal covariate shift.
\newblock In {\em International conference on machine learning}, pages
  448--456. PMLR, 2015.

\bibitem{isola2016imagetoimage}
Phillip Isola, Jun-Yan Zhu, Tinghui Zhou, and Alexei~A. Efros.
\newblock Image-to-image translation with conditional adversarial networks,
  2016.

\bibitem{7780682}
S.~D. {Jain} and K. {Grauman}.
\newblock Active image segmentation propagation.
\newblock In {\em 2016 IEEE Conference on Computer Vision and Pattern
  Recognition (CVPR)}, pages 2864--2873, 2016.

\bibitem{JohnsCVPR2015}
Edward Johns, Oisin Mac~Aodha, and Gabriel~J Brostow.
\newblock Becoming the expert-interactive multi-class machine teaching.
\newblock In {\em Proceedings of the IEEE Conference on Computer Vision and
  Pattern Recognition}, pages 2616--2624, 2015.

\bibitem{7299063}
C. {Kaeding}, A. {Freytag}, E. {Rodner}, P. {Bodesheim}, and J. {Denzler}.
\newblock Active learning and discovery of object categories in the presence of
  unnameable instances.
\newblock In {\em 2015 IEEE Conference on Computer Vision and Pattern
  Recognition CVPR}, pages 4343--4352, 2015.

\bibitem{kding2016active}
Christoph Kaeding, Erik Rodner, Alexander Freytag, and Joachim Denzler.
\newblock Active and continuous exploration with deep neural networks and
  expected model output changes, 2016.

\bibitem{kang2019contrastive}
Guoliang Kang, Lu Jiang, Yi Yang, and Alexander~G Hauptmann.
\newblock Contrastive adaptation network for unsupervised domain adaptation.
\newblock In {\em Proceedings of the IEEE/CVF Conference on Computer Vision and
  Pattern Recognition}, pages 4893--4902, 2019.

\bibitem{alex2019big}
Alexander Kolesnikov, Lucas Beyer, Xiaohua Zhai, Joan Puigcerver, Jessica Yung,
  Sylvain Gelly, and Neil Houlsby.
\newblock Big transfer (bit): General visual representation learning.
\newblock In Andrea Vedaldi, Horst Bischof, Thomas Brox, and Jan-Michael Frahm,
  editors, {\em Computer Vision -- ECCV 2020}, pages 491--507, Cham, 2020.
  Springer International Publishing.

\bibitem{LecunBackprop}
Yann~A LeCun, L{\'e}on Bottou, Genevieve~B Orr, and Klaus-Robert M{\"u}ller.
\newblock Efficient backprop.
\newblock In {\em Neural networks: Tricks of the trade}, pages 9--48. Springer,
  2012.

\bibitem{LEWIS1994148}
David~D. Lewis and Jason Catlett.
\newblock Heterogeneous uncertainty sampling for supervised learning.
\newblock In William~W. Cohen and Haym Hirsh, editors, {\em Machine Learning
  Proceedings 1994}, pages 148 -- 156. Morgan Kaufmann, San Francisco (CA),
  1994.

\bibitem{LI2018109}
Yanghao Li, Naiyan Wang, Jianping Shi, Xiaodi Hou, and Jiaying Liu.
\newblock Adaptive batch normalization for practical domain adaptation.
\newblock {\em Pattern Recognition}, 80:109 -- 117, 2018.

\bibitem{liu2016coupled}
Ming-Yu Liu and Oncel Tuzel.
\newblock Coupled generative adversarial networks.
\newblock In D. Lee, M. Sugiyama, U. Luxburg, I. Guyon, and R. Garnett,
  editors, {\em Advances in Neural Information Processing Systems}, volume~29.
  Curran Associates, Inc., 2016.

\bibitem{mishra2021surprisingly}
Samarth Mishra, Kate Saenko, and Venkatesh Saligrama.
\newblock Surprisingly simple semi-supervised domain adaptation with
  pretraining and consistency, 2021.

\bibitem{qiao2019weight}
Siyuan Qiao, Huiyu Wang, Chenxi Liu, Wei Shen, and Alan Yuille.
\newblock Weight standardization, 2019.

\bibitem{10.5555/1860625.1860629}
Piyush Rai, Avishek Saha, Hal Daume, and Suresh Venkatasubramanian.
\newblock Domain adaptation meets active learning.
\newblock In {\em Proceedings of the NAACL HLT 2010 Workshop on Active Learning
  for Natural Language Processing}, ALNLP 10, pages 27--32, USA, 2010.
  Association for Computational Linguistics.

\bibitem{romijnders2018domain}
Rob Romijnders, Panagiotis Meletis, and Gijs Dubbelman.
\newblock A domain agnostic normalization layer for unsupervised adversarial
  domain adaptation.
\newblock In {\em 2019 IEEE Winter Conference on Applications of Computer
  Vision (WACV)}, pages 1866--1875. IEEE, 2019.

\bibitem{10.5555645530.655646}
Nicholas Roy and Andrew McCallum.
\newblock Toward optimal active learning through sampling estimation of error
  reduction.
\newblock In {\em Proceedings of the Eighteenth International Conference on
  Machine Learning}, ICML 01, pages 441--448, San Francisco, CA, USA, 2001.
  Morgan Kaufmann Publishers Inc.

\bibitem{Rozantsev_2019}
Artem Rozantsev, Mathieu Salzmann, and Pascal Fua.
\newblock Beyond sharing weights for deep domain adaptation.
\newblock {\em IEEE Transactions on Pattern Analysis and Machine Intelligence},
  41(4):801--814, Apr 2019.

\bibitem{rukhovich2019mixmatch}
Danila Rukhovich and Danil Galeev.
\newblock Mixmatch domain adaptaion: Prize-winning solution for both tracks of
  visda 2019 challenge, 2019.

\bibitem{ILSVRC15}
Olga Russakovsky, Jia Deng, Hao Su, Jonathan Krause, Sanjeev Satheesh, Sean Ma,
  Zhiheng Huang, Andrej Karpathy, Aditya Khosla, Michael Bernstein,
  Alexander~C. Berg, and Li Fei-Fei.
\newblock Imagenet large scale visual recognition challenge.
\newblock {\em International Journal of Computer Vision (IJCV)},
  115(3):211--252, 2015.

\bibitem{10.55552034161.2034169}
Avishek Saha, Piyush Rai, Hal Daum\'{e}, Suresh Venkatasubramanian, and
  Scott~L. DuVall.
\newblock Active supervised domain adaptation.
\newblock In {\em Proceedings of the 2011 European Conference on Machine
  Learning and Knowledge Discovery in Databases - Volume Part III}, ECML PKDD
  11, pages 97--112, Berlin, Heidelberg, 2011. Springer-Verlag.

\bibitem{saito2019semisupervised}
Kuniaki Saito, Donghyun Kim, Stan Sclaroff, Trevor Darrell, and Kate Saenko.
\newblock Semi-supervised domain adaptation via minimax entropy.
\newblock In {\em Proceedings of the IEEE/CVF International Conference on
  Computer Vision}, pages 8050--8058, 2019.

\bibitem{10.2307/2331554}
Student.
\newblock The probable error of a mean.
\newblock {\em Biometrika}, 6(1):1--25, 1908.

\bibitem{su2019active}
Jong-Chyi Su, Yi-Hsuan Tsai, Kihyuk Sohn, Buyu Liu, Subhransu Maji, and
  Manmohan Chandraker.
\newblock Active adversarial domain adaptation.
\newblock In {\em Proceedings of the IEEE/CVF Winter Conference on Applications
  of Computer Vision}, pages 739--748, 2020.

\bibitem{sun2016deep}
Baochen Sun and Kate Saenko.
\newblock Deep coral: Correlation alignment for deep domain adaptation.
\newblock In {\em European conference on computer vision}, pages 443--450.
  Springer, 2016.

\bibitem{tan2019herbarium}
Kiat~Chuan Tan, Yulong Liu, Barbara Ambrose, Melissa Tulig, and Serge Belongie.
\newblock The herbarium challenge 2019 dataset, 2019.

\bibitem{tzeng2015simultaneous}
Eric Tzeng, Judy Hoffman, Trevor Darrell, and Kate Saenko.
\newblock Simultaneous deep transfer across domains and tasks.
\newblock In {\em Proceedings of the IEEE international conference on computer
  vision}, pages 4068--4076, 2015.

\bibitem{volpi2018generalizing}
Riccardo Volpi, Hongseok Namkoong, Ozan Sener, John Duchi, Vittorio Murino, and
  Silvio Savarese.
\newblock Generalizing to unseen domains via adversarial data augmentation.
\newblock In {\em Proceedings of the 32nd International Conference on Neural
  Information Processing Systems}, NIPS'18, page 5339–5349, Red Hook, NY,
  USA, 2018. Curran Associates Inc.

\bibitem{Wang09}
Josiah Wang, Katja Markert, and Mark Everingham.
\newblock Learning models for object recognition from natural language
  descriptions.
\newblock In {\em Proceedings of the British Machine Vision Conference}, 2009.

\bibitem{wang2018deep}
Mei Wang and Weihong Deng.
\newblock Deep visual domain adaptation: A survey.
\newblock {\em Neurocomputing}, 312:135--153, 2018.

\bibitem{NIPS2019_8470}
Ximei Wang, Ying Jin, Mingsheng Long, Jianmin Wang, and Michael~I Jordan.
\newblock Transferable normalization: Towards improving transferability of deep
  neural networks.
\newblock In H. Wallach, H. Larochelle, A. Beygelzimer, F. d~Alch~e Buc, E.
  Fox, and R. Garnett, editors, {\em Advances in Neural Information Processing
  Systems 32}, pages 1953--1963. Curran Associates, Inc., 2019.

\bibitem{wilson2018survey}
Garrett Wilson and Diane~J Cook.
\newblock A survey of unsupervised deep domain adaptation.
\newblock {\em ACM Transactions on Intelligent Systems and Technology (TIST)},
  11(5):1--46, 2020.

\bibitem{wu2018group}
Yuxin Wu and Kaiming He.
\newblock Group normalization.
\newblock In {\em Proceedings of the European conference on computer vision
  (ECCV)}, pages 3--19, 2018.

\bibitem{yang2017active}
Yazhou Yang and Marco Loog.
\newblock Active learning using uncertainty information.
\newblock In {\em 2016 23rd International Conference on Pattern Recognition
  (ICPR)}, pages 2646--2651. IEEE, 2016.

\bibitem{yoo2016pixellevel}
Donggeun Yoo, Namil Kim, Sunggyun Park, Anthony~S Paek, and In~So Kweon.
\newblock Pixel-level domain transfer.
\newblock In {\em European Conference on Computer Vision}, pages 517--532.
  Springer, 2016.

\bibitem{zhu2017unpaired}
Jun-Yan Zhu, Taesung Park, Phillip Isola, and Alexei~A Efros.
\newblock Unpaired image-to-image translation using cycle-consistent
  adversarial networks.
\newblock In {\em Proceedings of the IEEE international conference on computer
  vision}, pages 2223--2232, 2017.

\bibitem{Zhu03combiningactive}
Xiaojin Zhu, John Lafferty, and Zoubin Ghahramani.
\newblock Combining active learning and semi-supervised learning using gaussian
  fields and harmonic functions.
\newblock In {\em ICML 2003 workshop on The Continuum from Labeled to Unlabeled
  Data in Machine Learning and Data Mining}, pages 58--65, 2003.

\end{thebibliography}
	}
	
\end{document}


\begin{center}
		\textbf{\LARGE Supplementary Materials}
	\end{center}
	\setcounter{equation}{0}
	\setcounter{figure}{0}
	\setcounter{table}{0}
	\setcounter{page}{1}
	\makeatletter
	\renewcommand{\theequation}{S\arabic{equation}}
	\renewcommand{\thefigure}{S\arabic{figure}}

	\section{Experimental Setups}
	\subsection{Butterfly Classification}
	\begin{table}[h]
		\begin{center}
			\begin{tabular}{|c|c|c|c|}
				\hline 
				Hyperparameter/Classifier  & Source-only & Domain Adaptation & Target-only \\ 
				\hline 
				L2-Regularization & 0.01 & 0.05 & 0.01 \\ 
				\hline 
				Learning Rate & 0.001 & 0.0001 & 0.001 \\ 
				\hline 
			\end{tabular}
			\vspace*{0.4mm}
			\caption{This table shows the results of the grid search for the different setups.}
			\label{app:es}
		\end{center}
	\end{table} 

\subsection{Plant Classification}
\begin{table}[h]
	\begin{center}
		\begin{tabular}{|c|c|c|c|}
			\hline 
			Hyperparameter/Classifier  & Source-only & Domain Adaptation & Target-only \\ 
			\hline 
			L2-Regularization & 0.001 & 0.1 & 0.15 \\ 
			\hline 
			Learning Rate & 0.001 & 0.0001 & 0.0001 \\ 
			\hline 
		\end{tabular}
		\vspace*{0.4mm}
		\caption{This table shows the results of the grid search for the different setups.}
		\label{app:es}
	\end{center}
\end{table} 	
	\section{Further Examples from the Datasets}
	\subsection{Butterfly Dataset}
	\begin{figure}[h]
		\centering
		\includegraphics[scale=0.43]{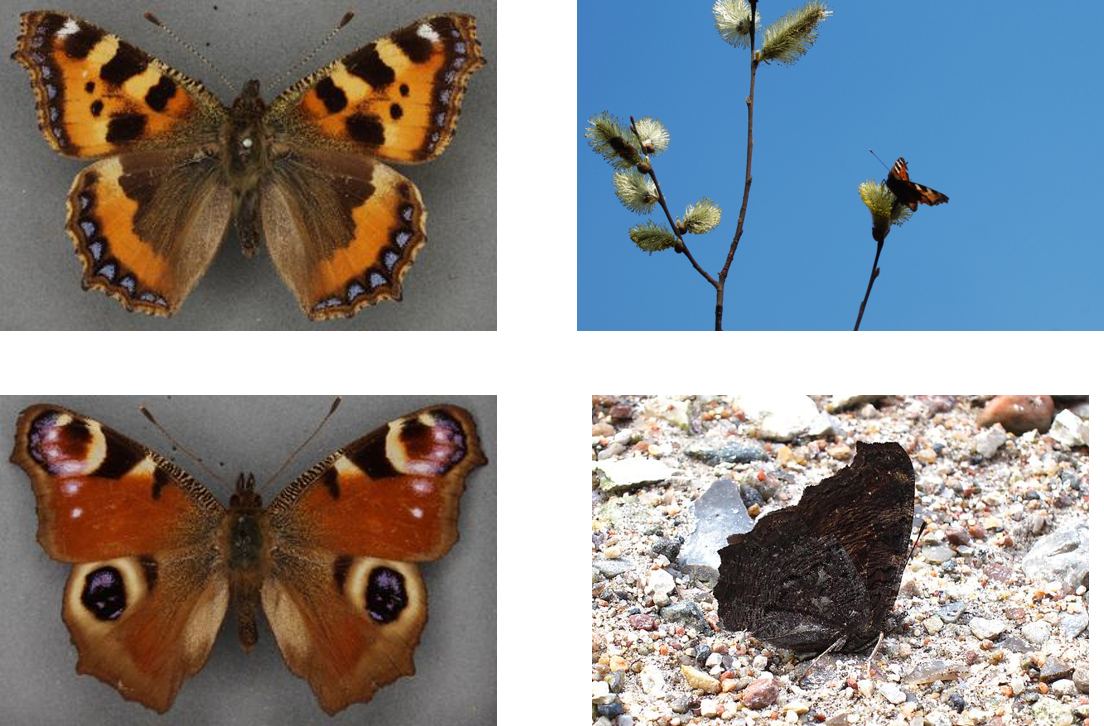}
		\caption[Difficult images in the butterfly dataset]{Here four images from the butterfly dataset are shown, which should illustrate possible difficulties in classification. On the left side images from the source domain and on the right side the corresponding images from the target domain can be seen.}
		\label{fig:ProbBut}
	\end{figure}
\newpage
\subsection{Plant Dataset}
\begin{figure}[h]
	\centering
	\includegraphics[scale=0.75]{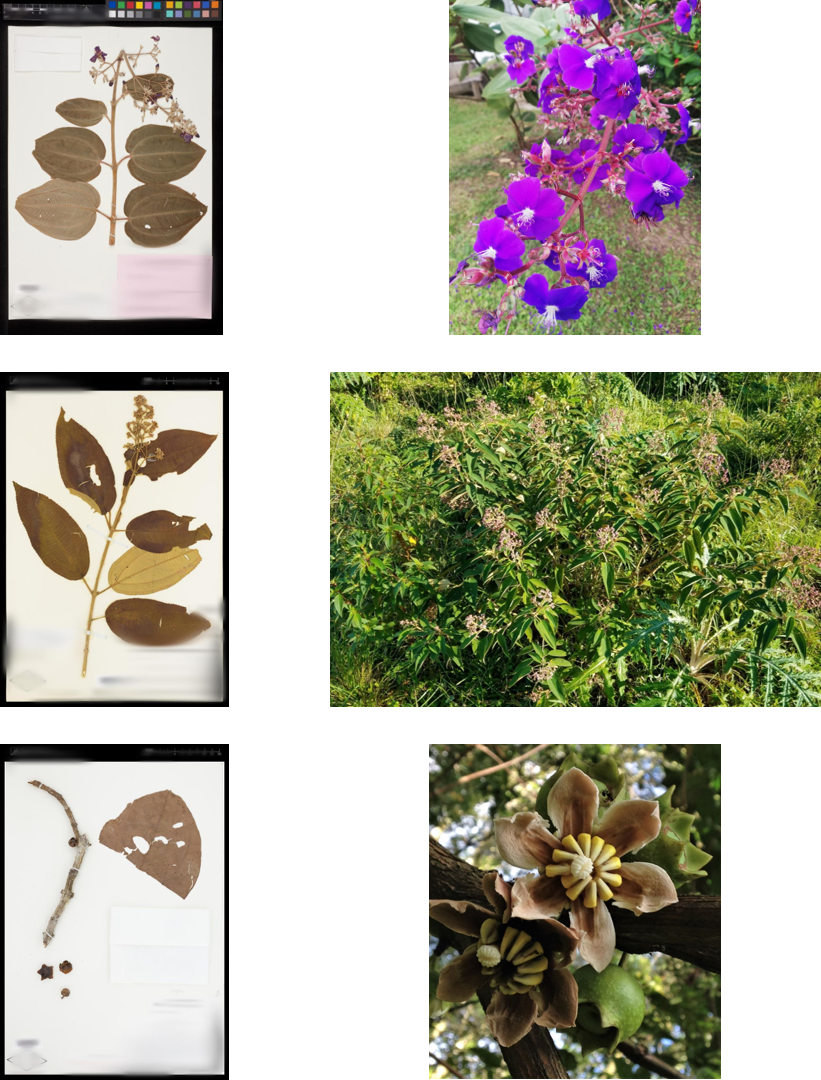}
	\caption[Difficult images in the plant dataset]{Here six images from the plant dataset are shown, which should illustrate possible difficulties in classification. On the left side images from the source domain and on the right side the corresponding images from the target domain are shown.}
	\label{fig:ProbPla}
\end{figure}

	\section{Results of the Significance Test}
	\label{app:exp3}
	\begin{landscape}
		\begin{table}[h]
			\begin{tabular}{|l|l|l|l|l|l|l|l|l|l|l|l|}
				\hline
				Round & 0 & 1 & 2 & 3 & 4 & 5 & 6 & 7 & 8 & 9 & 10 \\ \hline
				rnd - emoc & 0.66 & 0.56 & 0.85 & \textbf{0.03} & \textbf{0.03} & 0.1 & 0.13 & 0.12 & 0.5 & 0.27 & 0.07 \\ \hline
				rnd - iwerm & 0.05 & 0.91 & 0.18 & 0.28 & 0.25 & 0.71 & 0.37 & 0.3 & 0.61 & 0.48 & 0.14 \\\hline
				rnd - divdis & 0.25 & 0.9 & 0.31 & 0.7 & 0.59 & 0.76 & \textbf{0.01} & \textbf{0.03} & 0.28 & 0.15 & \textbf{0.02} \\\hline
				rnd - cer & 0.48 & 0.98 & 0.07 & 0.42 & 0.14 & 0.28 & 0.43 & \textbf{0.02} & 0.11 & \textbf{0.002} & \textbf{0.001} \\\hline
				iwerm - emoc & 0.07 & 0.52 & 0.1 & 0.13 & 0.21 & 0.23 & 0.29 & 0.38 & 0.29 & 0.46 & 0.3 \\\hline
				iwerm - divdis & 0.54 & 0.82 & 0.76 & 0.44 & 0.5 & 0.87 & \textbf{0.03} & 0.31 & 0.07 & 0.32 & 0.22 \\\hline
				iwerm - cer & 0.19 & 0.9 & 0.43 & 0.92 & 0.73 & 0.56 & 0.94 & 0.2 & \textbf{0.01} & \textbf{0.004} & \textbf{0.01} \\\hline
				divdis - emoc & 0.21 & 0.6 & 0.2 & 0.05 & 0.07 & 0.12 & 0.81 & 0.75 & 0.95 & 0.94 & 0.77 \\\hline
				divdis - cer & 0.57 & 0.94 & 0.33 & 0.61 & 0.31 & 0.34 & \textbf{0.04} & 0.6 & 0.56 & 0.17 & 0.27 \\\hline
				cer - emoc & 0.35 & 0.59 & \textbf{0.04} & 0.15 & 0.33 & 0.43 & 0.28 & 0.99 & 0.67 & 0.43 & 0.7 \\\hline
			\end{tabular}
			\caption[Results significance test for active learning]{This table presents the results of the student's t-test of the different active learning strategies on the plant dataset. Every combination of two strategies has been examined. If the calculated p-value is less than 0.05 the results are significantly different.}
			\label{tab:signi}
		\end{table}
	\end{landscape}